Authors:      Phillip Koshute, Jared Zook, Ian McCulloh

Title:        Recommending Training Set Sizes for Classification

Contact Information:

Phillip Koshute (corresponding author)
Johns Hopkins University Applied Physics Lab
11100 Johns Hopkins Road
Laurel, MD 20723
(240) 228-9546
Phillip.Koshute@jhuapl.edu

## ABSTRACT

Based on a comprehensive study of 20 established data sets, we recommend training set sizes for any classification data set. We obtain our recommendations by systematically withholding training data and developing models through five different classification methods for each resulting training set. Based on these results, we construct accuracy confidence intervals for each training set size and fit the lower bounds to inverse power low learning curves. We also estimate a sufficient training set size (STSS) for each data set based on established convergence criteria. We compare STSS to the data sets' characteristics; based on identified trends, we recommend training set sizes between 3000 and 30000 data points, according to a data set's number of classes and number of features. Because obtaining and preparing training data has non-negligible costs that are proportional to data set size, these results afford the potential opportunity for substantial savings for predictive modeling efforts.

## 1. INTRODUCTION

A key step in the development of a predictive model via supervised learning is to obtain and prepare data for training the model. To enable successful prediction, this training data must be of sufficient quality, adequately reflecting phenomena that are likely to arise when the model is applied. Additionally, the training data must also be of sufficient quantity. Without enough training data, underlying patterns are not identified by the learning algorithm and the model underperforms. At the same time, costs for obtaining, processing, and storing data can be extensive, both in terms of computational resources and human effort. Therefore, for many applications, there is strong motivation to identify the "right" amount of data.

Stemming from this motivation, there is one question that is consistently asked at the outset of almost any predictive modeling effort: How much data is enough? I.e., what training set sizes are sufficient? In this paper, we begin to answer this question, providing concrete recommendations on appropriate training set sizes for predictive model development.

Our scope specifically involves supervised classification problems. To reach our recommendations, we studied twenty benchmark classification data sets, systematically withholding training data from each and evaluating the effect of these training set reductions on performance of five prominent supervised classification methods. With each set's results, we fitted a learning curve, estimating model performance for every possible training set size.

Based on these fitted curves, we identified a sufficient training set size (STSS) for each set. We defined the STSS as the minimum training set size for which projected accuracy is within 5% of the best achievable accuracy with at least 90% confidence. Comparable convergence criteria is used by John and Langley [1] and Provost et al. [2].

Seeking to generalize our results, we linked variations in STSS for different data sets to measurable characteristics of those sets. The characteristics that we studied included number of features, proportion of features that are categorical, and number of classes (cf., Section 3.1). Based on these relationships, we propose a general framework for determining the STSS of any data set that might be encountered. Figure 1 summarizes our approach.

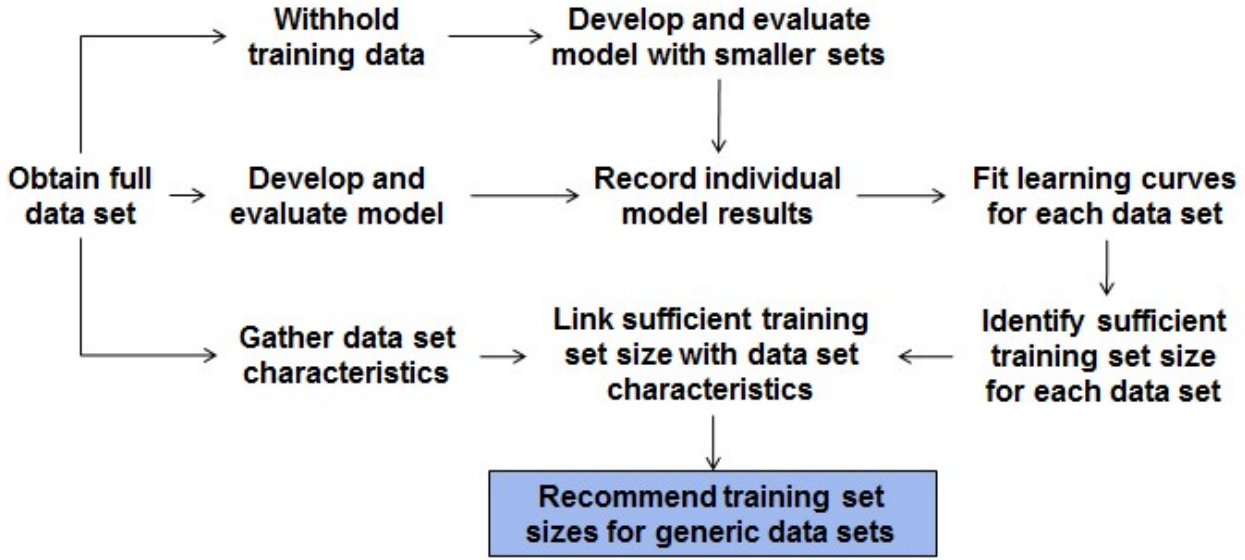

**Figure 1: Process for Recommending Training Set Sizes**

The remainder of this report is organized as follows. Section 2 reviews others' related work and perspectives. Section 3 describes the studied data and the machine learning methods. Section 4 presents the results of these initial modeling efforts, as well the estimated learning curves. Section 5 highlights the STSS for each set and details the recommendations for unseen data sets. Section 6 discusses these results and suggests potential avenues for further study.

## 2. RELATED WORK

Within classical statistics, there are well-established methods for determining minimum sample sizes that are necessary to achieve certain error rates (cf., [3]). Using such techniques, a learning curve could be inferred by plotting minimum sample sizes against corresponding error rates. These techniques are known almost universally among statisticians. However, comparable extensions into machine learning studies are relatively few and have yielded only modest and nuanced results.

Similarly, sampling theory has been invoked to estimate STSS according to the formula for determining the sample size required to estimate the mean value of a population with certain precision. The formula for this approach is $N_S = Z^2\sigma^2/h^2$, where $\sigma$ is the presumed standard deviation of the distribution, $Z$ is a measure of confidence, and $h$ is the desired confidence interval half-width [4, 5]. However, since $\sigma$, $Z$, and $h$ are not straightforward to define for classification problems, this formula offers more theoretical foundation than practical guidance.

Within studies focused on predictive models, a common technique to estimate for STSS for a given data set is to measure classification accuracy for smaller training set sizes and use these data points to estimate the accuracy for every possible training set size. The resulting model is called a "learning curve." As a colloquial term, learning curve refers to how a person's skill in a given area changes (and generally improves) as more experience is received. In predictive model development, the "experience" is training data. Once a learning curve has been estimated, the necessary training set size to achieve certain accuracy can be calculated.

Learning curve approaches have been applied to natural language processing [6], medical imaging [7, 8], and sampling design [1, 2]. Others have implicitly analyzed learning curves in considering the effect of training set size on classifier performance; e.g., [9], [10], and [11]. Figueroa et al. [12] propose a general method for estimating learning curves, demonstrating its capability with both patient categorization and waveform discrimination. A notional learning curve is shown in Figure 2.

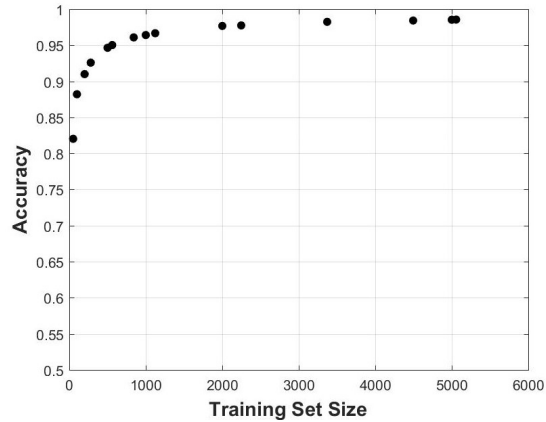

**Figure 2: Learning Curve Example**

Others have proposed methods for identifying STSS for particular domains including cancer screening [13, 14], fault detection [15], land cover classification [16], and social media analysis [17]. Similarly, Baum and Haussler [18] propose a method for identifying STSS for neural networks.

However, to our knowledge, no study has attempted to extend their results to recommending STSS for unseen data sets, either with learning curves or other methods.

## 3. METHODOLOGY

As introduced in Section 1, we incorporated a variety of data sets and methods in our study of STSS. For each selected data set (Section 3.1), we developed models with various supervised learning classification methods (Section 3.2) using different sizes of training sets (Section 3.3). Based on these preliminary model results, we fitted learning curves that model the best achievable classification accuracy with a given training set (Section 3.4). These learning curves enable us to estimate STSS for each data set (Section 3.5).

## 3.1 DATA OVERVIEW

We selected twenty data sets from the UC-Irvine Machine Learning Repository [19]. Because of the prominence of this repository, we regard these data sets as "gold standard." They have wide range of sizes and applications, as shown in Table 1. Each data set was denoted with a three-letter code.

**Table 1: Data Set Summary**

| Classification Data Set | Code | # Data Points ($N_P$) | # Features ($N_F$) |
|---|---|---|---|
| Annealing [20] | ANN | 798 | 38 |
| Breast Cancer [21] | BCW | 683 | 9 |
| Bank Marketing [22] | BMK | 41188 | 20 |
| Cardiotocography [23] | CAR | 2126 | 21 |
| Car Evaluation [24] | CEV | 1728 | 6 |
| Congressional Voting Records [25] | CVR | 435 | 16 |
| Credit Card Defaults [26] | DCC | 30000 | 23 |
| Diabetic Retinopathy [27] | DRB | 1151 | 19 |
| Handwritten Digit Recognition [28] | DRE | 5619 | 64 |
| Indian Liver Patient [29] | ILP | 583 | 10 |
| Ionosphere [30] | ION | 351 | 34 |
| Letter Recognition [31] | LRE | 20000 | 16 |
| Gamma Telescope Particles [32] | MAG | 19020 | 10 |
| Mice Protein Expression [33] | MPE | 972 | 80 |
| Mushroom Edibility [34] | MSH | 8124 | 22 |
| Occupancy Detection [35] | OCD | 20560 | 5 |
| Phishing Websites [36] | PHW | 11044 | 30 |
| QSAR Biodegradability [37] | QBD | 1055 | 41 |
| Skin Segmentation [38] | SSG | 245057 | 3 |
| Wall-Following Robots [39] | WFR | 5456 | 24 |

Several data processing steps were taken for each set. Categorical variables were converted to sets of binary variables. For instance, if a given variable $X$ had the possible values {*red*, *green*, *blue*}, it would be replaced with three binary variable $X_{red}$, $X_{green}$, and $X_{blue}$, which each had the possible values {0, 1}. The number of features, shown in Table 1, was counted before this conversion. Continuous variables were standardized, such that each had a mean of 0 and standard deviation of 1.

Other processing steps were taken for missing values. For categorical variables, missing values were treated as a separate category. Continuous features that had missing values for more than 10% of all points were removed entirely. Similarly, data points with missing values for continuous features were removed. The number of data points, shown in Table 1, was counted after these points were removed.

We also gathered several other characteristics of each data set, including number of categorical features ($N_{CAT}$), number of continuous features ($N_{CONT}$), proportion of features that are categorical ($R_{CAT}$), number of target classes ($N_C$), minimum single-class proportion ($C_{MIN}$), and class imbalance ($I_C$). The latter is a measure of the difference in actual and expected proportion of the least common target class ($L$) and its expected value ($L_E$); i.e., $I = (L_E - L)/L_E$. These values are shown in Table 2.

**Table 2: Additional Data Set Characteristics**

| Set | $N_{CAT}$ | $N_{CONT}$ | $R_{CAT}$ | $N_C$ | $C_{MIN}$ | $I_C$ |
|-----|-----------|------------|-----------|-------|-----------|-------|
| ANN | 32 | 6 | 0.842 | 5 | 0.001 | 0.956 |
| BCW | 0 | 9 | 0.000 | 2 | 0.350 | 0.300 |
| BMK | 10 | 10 | 0.500 | 2 | 0.113 | 0.775 |
| CAR | 0 | 21 | 0.000 | 3 | 0.083 | 0.752 |
| CVE | 6 | 0 | 1.000 | 4 | 0.038 | 0.850 |
| CVR | 16 | 0 | 1.000 | 2 | 0.386 | 0.228 |
| DCC | 3 | 20 | 0.130 | 2 | 0.221 | 0.558 |
| DRB | 0 | 19 | 0.000 | 2 | 0.469 | 0.062 |
| DRE | 0 | 64 | 0.000 | 10 | 0.098 | 0.016 |
| ILP | 1 | 9 | 0.100 | 2 | 0.286 | 0.427 |
| ION | 0 | 34 | 0.000 | 2 | 0.359 | 0.282 |
| LRE | 0 | 16 | 0.000 | 26 | 0.037 | 0.046 |
| MAG | 0 | 10 | 0.000 | 2 | 0.322 | 0.357 |
| MPE | 3 | 77 | 0.038 | 8 | 0.092 | 0.259 |
| MSH | 22 | 0 | 1.000 | 2 | 0.482 | 0.036 |
| OCD | 0 | 5 | 0.000 | 2 | 0.231 | 0.538 |
| PHW | 30 | 0 | 1.000 | 2 | 0.443 | 0.114 |
| QBD | 0 | 41 | 0.000 | 2 | 0.337 | 0.325 |
| SSG | 0 | 3 | 0.000 | 2 | 0.208 | 0.585 |
| WFR | 0 | 24 | 0.000 | 4 | 0.060 | 0.760 |

## 3.2 MACHINE LEARNING METHODS

We implemented five supervised learning classification methods: logistic regression, naïve Bayes, neural networks, random forests, and support vector machines. These methods were selected to span the range of most common classification approaches within machine learning. By developing models with these methods, we intended to represent a fairly substantial modeling effort by a typical machine learning expert. To develop models, we implemented appropriate classifiers in Python [40], always with default settings unless otherwise noted.

Logistic regression involves calculating a probability for each target class $y_i$, according to the logistic equation $P(y_i|x) = 1/(1+e^{-\beta_i x})$, where $\beta_i$ are the learned coefficients for class $y_i$. A classification is made by identifying the most probable class. For this method, we the implemented 'LogisticRegression' classifier in Python.

Naïve Bayes similarly involves calculating a probability for each $y_i$, based on the conditional relationships between each target class and a given value of a feature $x_k$, $P(y_i|x) = \Pi_k P(y_i|x_k)$. As with logistic regression, a classification is made by identifying the most probable class. With naïve Bayes, all of the features are assumed to be independent of each other. Because all of our features are continuous or binary, we implemented the 'GaussianNB' classifier in Python.

Neural networks involve layers of nodes, in which each layer's values are functions of the previous layer's nodes. The network's input nodes are comprised of feature values and it has one output node for each target class. A classification is made by selecting the class corresponding to the output node with the highest value. We implemented the 'MLPClassifier' classifier in Python, specifying one hidden layer and $K*N_F$ hidden nodes, where $N_F$ is a data set's number of features and $K$ is a specified constant. We used $K = 3$ and verified that performance was generally not affected by this selection (cf., Section 4.2.2).

Random forests are ensembles of decision trees that specify logical rules based on feature values to assign a class. The decision trees are constructed from various subsets of the original training set. A classification is made by selecting the class that has been selected by the most individual decision trees within the random forest. We implemented the 'RandomForestClassifier' classifier in Python, specifying 100 trees in each model.

Support vector machines (SVM) identify hyperplanes with a data set's feature space that optimally separate data points corresponding to two classes. This method can also be applied to data sets with more than two target classes by nesting one-vs-all classifiers. However, due to computational limitations, we opted to only develop support vector models for two-class data sets with less than 10,000 data points. We implemented the 'SVC' classifier in Python.

Table 3 summarizes the methods under consideration.

**Table 3: Implemented Supervised Learning Classification Methods**

| Method | Python Implementation | Notes |
|---|---|---|
| Logistic Regression | 'LogisticRegression' | --- |
| Naïve Bayes | 'GaussianNB' | --- |
| Neural Networks | 'MLPClassifier' | # hidden layers = 1; # hidden nodes = 3*(# features) |
| Random Forests | 'RandomForestClassifier' | # trees = 100 |
| SVM | 'SVC' | Only run with two-class data sets |

## 3.3  TRAINING SET SIZES

For each selected data set, we developed 4000 or 5000 models according to the following Monte Carlo, cross-validation scheme. (We did not develop SVM models for data sets with more than two classes or with more than 10000 data points.) We performed ten-fold cross-validation and used *StratifiedKFold* in Python was used to evenly distribute each target value among all folds. From each full training set, we randomly selected subsets of different sizes that are used to train additional models that can be evaluated against the full test fold.

The training set sizes for a given data set with full size $N$ were the union of the following:

- $A = \{0.05N, 0.10N, 0.15N, 0.20N, 0.40N, 0.60N, 0.80N, 0.90N\}$
- $B_j = 5*10^j$, for all positive $j$ such that $B_j \leq 0.90N$
- $C_j = 10^{j+1}$, for all positive $j$ such that $C_j \leq 0.90N$
- $D_j = 2*10^{j+1}$, for all positive $j$ such that $D_j \leq 0.90N$

Following this scheme, between 11 and 20 training set sizes were evaluated for each data set (depending on $N_P$ for the particular data set). From among each set of training folds, ten subsets were randomly sampled from the training folds. Since each data set was separated into ten cross-validation folds, there are 100 training sets for each training set size for each data set. With each method (described in Section 3.2), we developed a model with each training set.

## 3.4 LEARNING CURVE FITTING

For each training set, we selected the best accuracy value from among the implemented models and constructed a 80% confidence interval from among the best values of each of the 100 training sets. It is possible that different methods were represented in the set of best results for any particular data set and training set size. We determined confidence intervals by an exact method, setting the upper bound as the 90[th] highest accuracy value and the lower bound as the 11[th] highest accuracy value.

We gathered lower bounds of the confidence intervals for each data set and training set size. The resulting plot of accuracy lower bounds and training set sizes yields an empirical learning curve for each data set.

To account for noise, we replaced each empirical learning curve by a fitted learning curve, using the inverse power law method from Figueroa et al. [12]. This method fits a curve according to $y = f(x) = \alpha - \beta x^\gamma$, where $y$ is classification accuracy, $x$ is training set size, and $\alpha$, $\beta$, and $\gamma$ are curve-fitting parameters.

We learned the separate parameters for each data set using MATLAB's *fmincon* function, setting $\{\alpha = 1, \ \beta = 0, \ \gamma = 0\}$ as initial parameter values. To sensibly guide the optimization search, we provided the following constraints to the model, where $y_{max}$ is the best achieved classification accuracy with any training set size: $y_{max} \leq \alpha \leq 1; \beta \geq 0; \gamma \leq 0$. Within *fmincon*, we used the *sqp* algorithm.

## 3.5 SUFFICIENT TRAINING SET SIZE CRITERIA

To identify a STSS for each set, we found the minimum training set size $x$ such that $P(f(\infty) - f(x) < T_2) \geq T_1$, as used by [1] and [2], where $f$ is the fitted learning curve. This criteria requires that the model performance have probability $T_1$ of being within $T_2$ of the best possible performance. We set $T_1 = 0.9$, which corresponds to the interval above the lower bound of the 80% confidence interval, and $T_2 = 0.05$. For each set, we estimated $f(\infty)$ as the value of the coefficient $\alpha$ for a learning curve fit to the mean classification accuracy at each training set size according to the same fitting process described in Section 3.4.

In practice, we calculated $STSS = \exp((1/\gamma)*(\log(\alpha - f(\infty) + T_2) - \log(\beta)))$, which is the minimum training set size $x$ that satisfies $f(\infty) - T_2 = \alpha - \beta x^\gamma$, where $\alpha$, $\beta$, and $\gamma$ characterize the learning curve (cf., Section 3.4) fitted to the lower bound values of the 80% confidence interval of the preliminary model results.

## 4. RESULTS

We first gathered the results of preliminary models, enabling the construction of empirical learning curves for each data set. Once we were relatively assured that these results were realistic, we fitted these results to inverse power law learning curves.

## 4.1 PRELIMINARY MODEL RESULTS

The mean values and confidence intervals for each data set and training set size are plotted in Figure 3. These results comprise each data set's empirical learning curve. With each data set, performance generally improves as more training data becomes available, but the rate of improvement (measured as increase in accuracy for each additional training point) diminishes with larger training sets. This trend is consistent with the general trend of an inverse power law model, which we fit in Section 4.3.

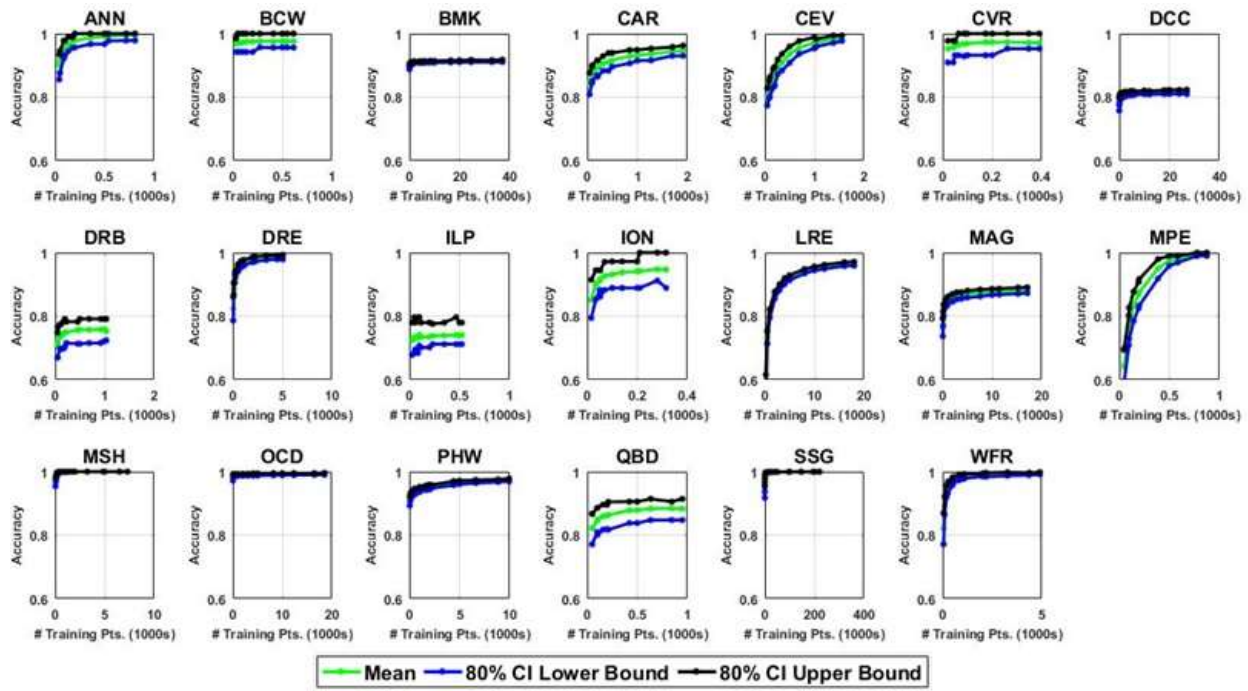

**Figure 3: Empirical Learning Curves**

## 4.2  ROBUSTNESS CHECKS

Since the results of the preliminary models form the foundation for our subsequent analysis, it is important to have reasonable assurance that these results are fairly close to the best achievable performance. We validated these results in three ways: by comparing them with others' published results on the same data sets, when possible; by evaluating the potential benefit of parameter tuning; and by evaluating the potential benefit of feature selection. Due to computational and time limitations, we did not incorporate parameter tuning or feature selection into our preliminary model development.

## 4.2.1 COMPARISON WITH OTHERS' RESULTS

For many of our selected data sets, others have developed classification models and published their classification accuracy results. While some of the authors have evaluated their models in slightly different ways (e.g., different training-test data splits or cross-validation schemes), these comparisons generally indicate whether our results are competitive. Table 4 shows these comparisons. In some cases, we did not find any published results for a given data set or the published results did not include classification accuracy.

For each case in which others' results were available for comparison, our models performed at least as well, and in several cases, our models performed better.

**Table 4: Comparison of Achieved Accuracy with Others' Published Accuracy for Each Data Set**

| Data Set | Published Accuracy | Mean Achieved Accuracy | Paper |
|---|---|---|---|
| ANN | 0.93 | 1.00 | Jiang 2004 |
| BCW | 0.97 | 0.98 | Wolberg 1994 |
| BMK | Accuracy not reported | 0.91 | Moro 2014 |
| CAR | 0.86 | 0.95 | Ayres-de-campos 2000 |
| CEV | 0.95 | 0.99 | Oza and Russell 2001 |
| CVR | 0.96 | 0.97 | Bonet 1998 |
| DCC | 0.82 | 0.82 | Yeh 2009 |
| DRB | No paper found | 0.76 | --- |
| DRE | 0.94 | 0.99 | Xu 1992 |
| ILP | Accuracy not reported | 0.74 | Ramana 2011 |
| ION | Accuracy not comparable | 0.95 | Raymer, et al. 2003 |
| LRE | 0.83 | 0.97 | Frey 1991 |
| MAG | 0.56 | 0.88 | Savicky and Kotrc 2004 |
| MPE | Accuracy not reported | 1.00 | Higuera 2015 |
| MSH | 0.99 | 1.00 | Duch 1997 |
| OCD | 0.99 | 0.99 | Candanedo 2016 |
| PHW | 0.94 | 0.97 | Mohammad 2014 |
| QBD | 0.83 | 0.88 | Mansouri 2013 |
| SSG | 0.94 | 1.00 | Bhatt 2009 |
| WFR | 0.97 | 0.99 | Freire 2009 |

## 4.2.2 EFFECT OF PARAMETER TUNING

The success of a supervised learning method requires a proper setup, e.g., selecting appropriate parameter values. Our preliminary modeling methods require various decisions: for neural networks, numbers of hidden layers and of hidden nodes; number of trees for random forests; kernel type for support vector machines. Aside from our values for number of hidden nodes (described in Section 3.2), we generally utilized the default options for the respective Python classifier. However, to gain an initial sense of potential benefit of parameter tuning, we conducted brief experiments in which we incorporated parameter tuning for the number of hidden nodes in a neural network.

We developed neural network models for the CAR, DRB, and WFR data sets that included an additional parameter tuning step. We selected these data sets because they are comprised of a moderate amount of data and had underperforming others' neural network results. For these models, we separated each randomly selected training data set into separate training and validation sets. A quarter of the available training data was set aside as a validation set; the remaining training data was used to develop models with various values of $K$, while the validation set was used to evaluate each of these models and identify the optimal value of $K$. For comparison, we also developed models with all available training data without any parameter tuning. The remaining test fold was used to evaluate both models.

Figure 4 compares the results on these data sets with and without parameter tuning for the three selected sets. None of the sets show any improvement with parameter tuning, suggesting that our decision to not tune parameters did not negatively affect preliminary model results. For many training set sizes, the models developed with parameter tuning perform slightly worse than models developed without parameter tuning (against the test set). With identical training sets, this seems counterintuitive, since the tuned parameters will take on the default value if no other value yields better performance (against the validation set). However, parameter tuning ultimately requires reducing the training set, in favor of keeping some data as a validation set. In these cases, it seems as though the benefits of parameter tuning are outweighed by the drawbacks of reducing the training set.
.

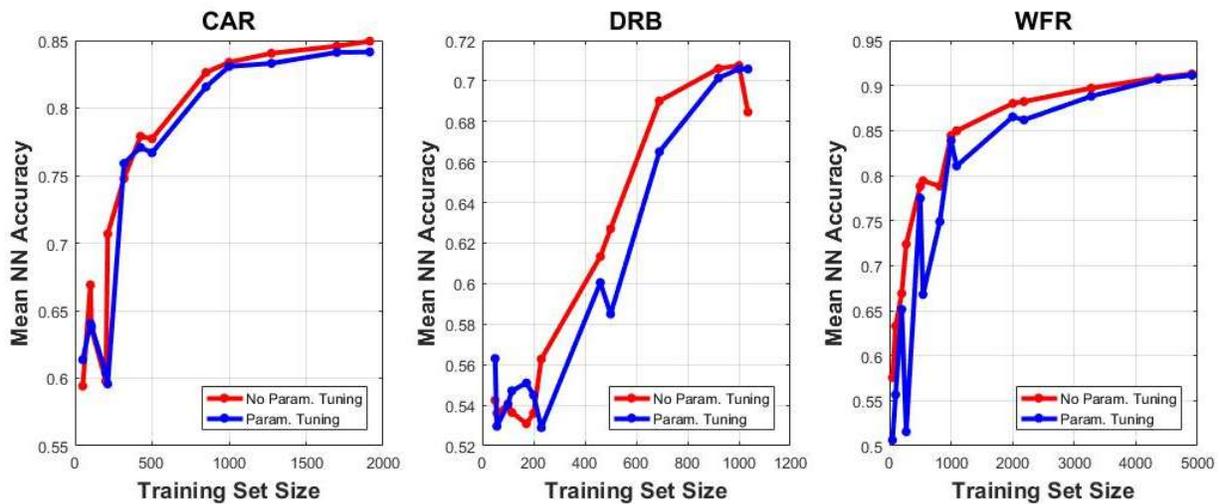

**Figure 4: Comparison of Selected Neural Network Models With and Without Parameter Tuning**

## 4.2.3 EFFECT OF FEATURE SELECTION

Feature selection is another important step in model development. Keeping superfluous features can create noise and prevent underlying patterns from being discovered. Discarding important features can eliminate otherwise helpful detail. In the development of our preliminary models, we assumed that the data sets' features were generally optimal and that the classification methods would suffice for any final down-selecting. (Methods implicitly "select" features by giving lower weights to those features.) Since we obtained our data from an established and widely accepted repository [19], this seemed to be a reasonable assumption. Nevertheless, similar to parameter tuning, we sought to also evaluate the effect of this decision.

As with parameter tuning, we developed parallel models (with and without feature selection) for a subset of data sets. Since feature selection is applicable to any method, we developed these models for all applicable methods. We evaluated the DRB, ILP, PHW, and QBD data sets, selecting them because their performance was generally poorer than for other data sets. For each pair of data set and method, we identified the best classification accuracy in the models developed with and without feature selection. Table 5 compares these results, which suggest that feature selection may have a substantial effect on model improvement; on average, incorporating feature selection increases classification accuracy by 4.5%. In future efforts, incorporating feature selection in preliminary modeling would seem to be warranted.

**Table 5: Mean Change in Best Classification Accuracy with Feature Selection**

| Data Set | Change with Feature Selection |
|----------|-------------------------------|
| DRB | +4.9% |
| ILP | +6.5% |
| PHW | +1.3% |
| QBD | +5.4% |

## 4.3  INVERSE POWER LAW LEARNING CURVES

Having validated the preliminary model results, we fitted inverse power law curves to the lower bound of the 80% confidence interval values for each data set. We considered the confidence interval lower bounds in order to have reasonable assurance that a given accuracy level could be achieved and to more easily apply the STSS criteria described in Section 3.5. This method generally fits the corresponding empirical learning curves very well; the MAE vary from 0.0005 to 0.0245 with a median of 0.0027. Figure 5 plots the empirical and fitted learning curves.

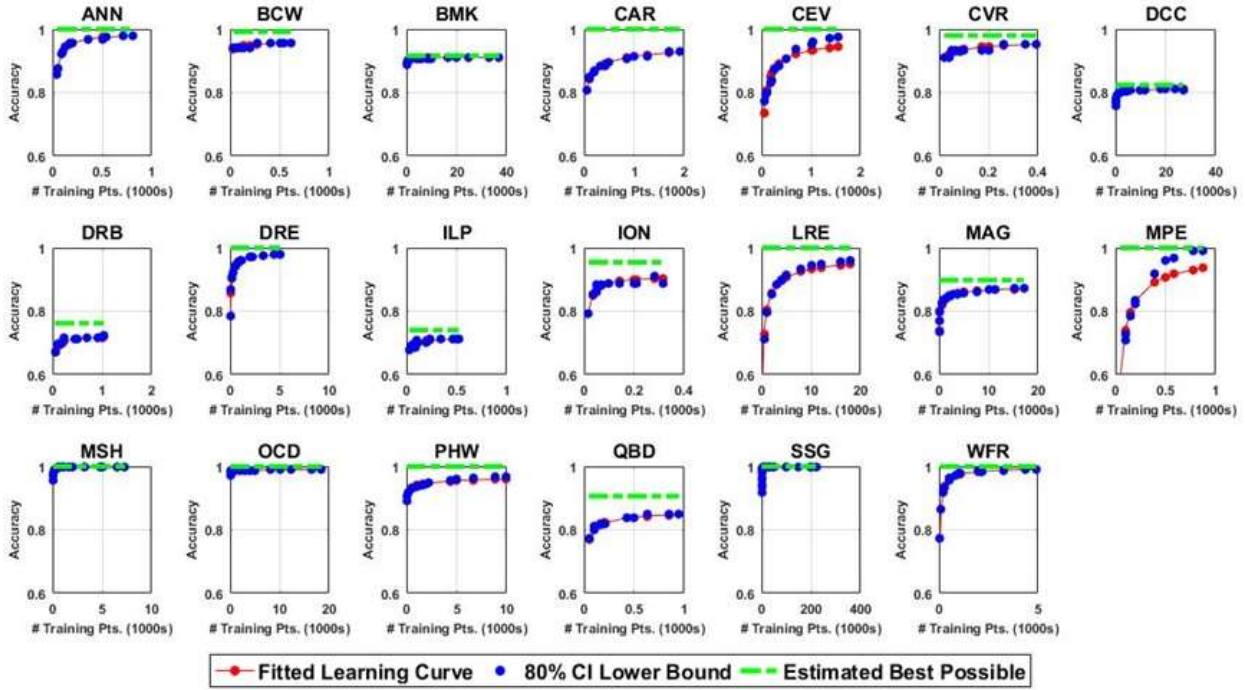

**Figure 5: Fitted Inverse Power Law Learning Curves**

## 5. SUFFICIENT TRAINING SET SIZES

As described in Section 3.5, we calculated a STSS for each data set based on these fitted curves, finding the minimum training set size for which the fitted classification accuracy is within 5% of the estimated best possible accuracy. These values are shown in Table 6.

For some data sets in which model performance converged very quickly, the calculated STSS is less than the minimum training set size included in the preliminary models, $x_{min}$. In such cases, the actual STSS lies within $[1, x_{min}]$ and consequently is not known precisely. For such data sets, we conservatively estimated $STSS = x_{min}$. These sets are indicated with an asterisk in Table 6.

**Table 6: Calculated Sufficient Training Set Size (STSS)**

| Data Set | STSS | | Data Set | STSS |
|----------|------|---|----------|------|
| ANN | 161 | | ION | 329 |
| BCW | 65 | | LRE | 18944 |
| BMK* | 50 | | MAG | 2044 |
| CAR | 7598 | | MPE | 1290 |
| CEV | 1891 | | MSH* | 50 |
| CVR | 69 | | OCD* | 50 |
| DCC | 118 | | PHW | 2851 |
| DRB | 404 | | QBD | 1844 |
| DRE | 677 | | SSG | 113 |
| ILP | 50 | | WFR | 382 |

## 5.1  ANALYSIS OF MEASURABLE DATA SET CHARACTERISTICS

In attempt to understand whether STSS might be anticipated, we compared data set characteristics (cf., Section 3.1) with calculated STSS values. Specifically, we used MATLAB's *stepwisefit* tool, which constructs a linear regression model, using forward stepwise feature selection. We set the *p*-value thresholds to add or remove a feature from the model both at 0.05. The results of this analysis, which are shown in Table 7, provide an indication of whether a given characteristic is predictive of STSS. These results suggest that STSS is generally higher for data sets with more classes and fewer features.

**Table 7: Comparison of Data Set Characteristics and Sufficient Training Set Size (STSS)**

| Feature | Selected in a Linear Model | Sign of the Coefficient |
|---------|---------------------------|-------------------------|
| $N_{CAT}$ | No | n/a |
| $N_{CONT}$ | No | n/a |
| $N_F$ | Yes ($p = 0.038$) | Negative |
| $R_{CAT}$ | No | n/a |
| $N_C$ | Yes ($p < 0.001$) | Positive |
| $C_{MIN}$ | No | n/a |
| $I_C$ | No | n/a |

## 5.2  TRAINING SET SIZE RECOMMENDATIONS

To extend the analysis of data set characteristics to training set size recommendations, we grouped each data set, based on its number of classes and number of features (the characteristics selected in Section 5.1). We established two bins for each characteristic, separating the data sets based on the median value for the characteristic; the median number of classes was two and the median number of features was 20. The resulting groups of data sets are described in Table 8 and further illustrated in Figure 6.

**Table 8: Number of Data Sets and Maximum Sufficient Training Set Size (STSS) in Each Data Set Group**

|  |  | Number of Classes | |
|---|---|---|---|
|  |  | $N_C = 2$ | $N_C > 2$ |
| Number of Features | $N_F \leq 20$ | 8 data sets; Max. STSS = 2044 | 2 data sets; Max. STSS = 18944 |
|  | $N_F > 20$ | 5 data sets; Max. STSS = 2851 | 5 data sets; Max. STSS = 7598 |

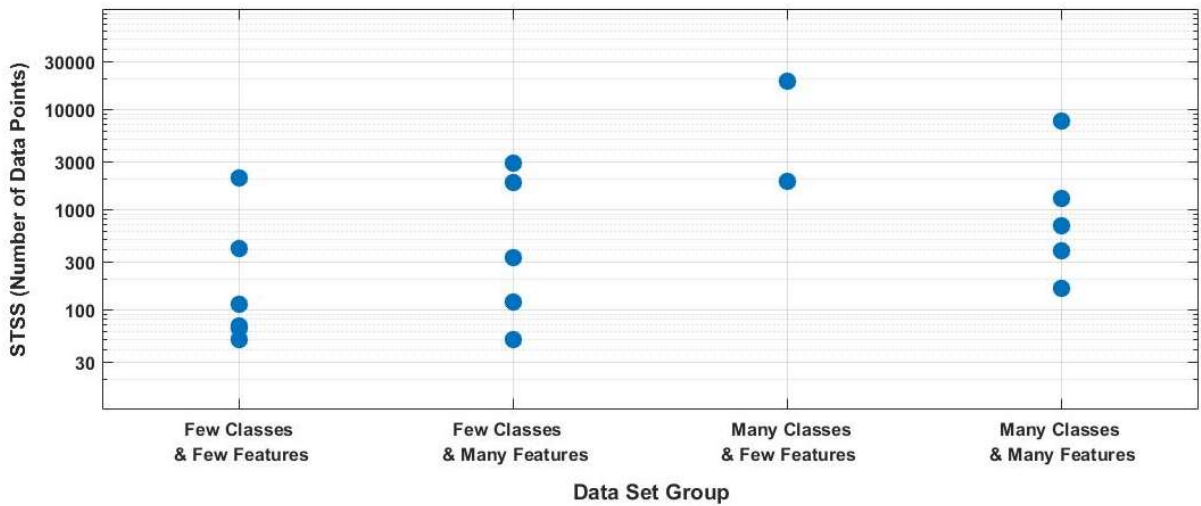

**Figure 6: Sufficient Training Set Size (STSS) by Data Set Group**

To obtain recommendations for training set sizes for a given data set, we rounded up the maximum STSS in each data set group to the nearest $10^M$ or $3 \times 10^M$; this is approximately equal to the nearest half-order of maximum, as can be seen in Figure 10. These conventions yield the following training set sizes recommendations:

- For sets with more than 2 classes and 20 or fewer features, 30000 training points are needed
- For sets more than 2 classes and more than 20 features, 10000 training points are needed
- For sets with 2 classes and any number of features, 3000 training points are needed

Though these recommendations have been obtained from a specific collection of data sets and are necessarily preliminary, they offer initial guidance for any unseen data set for which a supervised classification model is sought.

# 6. DISCUSSION AND NEXT STEPS

Based on a comprehensive study of 20 established data sets, we have recommended training set sizes for any classification data set. To our knowledge, our study is the first of its kind to make such recommendations with this level of generality. Even at the beginning of a project, if an analyst can estimate the number of classes and number of features in their prospective data set, they can apply our recommendations. Because obtaining and preparing training data has non-negligible costs that are proportional to data set size, these results afford the potential opportunity for substantial savings for predictive modeling efforts.

We anticipate several avenues for extending our analysis:

- Our underlying preliminary model results may be improved by incorporating additional data sets, including larger data sets; feature selection; or additional classification methods, e.g., Bayesian networks.

- Data sets may be characterized in additional ways. For instance, codifying data sets according to their domain, such as image processing or cybersecurity, may highlight nuances in training set size requirements. Alternatively, data sets might be grouped according to how their target classes have been assigned; more training data may needed when target classes have been manually assigned or in other cases that error tendencies are higher.

- Seeking to make our recommendations as general as possible, we have considered only the method that performs best with a given training set. In some cases, though, it might beneficial to have recommended training set sizes for specific methods. Our analysis could be repeated accordingly.

- Though our efforts focused on classification problems, they could seemingly be repeated with regression problems. Based on learning curves for a representative set of regression sets, sufficient training set sizes could be generally recommended for other regression data sets.

# ACKNOWLEDGMENTS

This work was funded by the following organizations: Johns Hopkins University Applied Physics Laboratory; Department of the Navy, Office of Naval Research.

# DISCLAIMER

Any opinions, findings, and conclusions or recommendations expressed in this material are those of the author(s) and do not necessarily reflect the views of the Office of Naval Research.